\documentclass[sigconf,nonacm]{acmart}

\usepackage[noend]{algpseudocode}
\usepackage{algorithm}
\makeatletter
\def\algbackskip{\hskip-\ALG@thistlm}
\makeatother
\usepackage{bbm}
\usepackage{comment}
\usepackage{tikz}
\usepackage{booktabs}
\usetikzlibrary{positioning}
\usepackage{bm}
\usepackage{pgfplots}
\usepackage{paralist}
\usepackage{caption}
\usepackage{subcaption}
\usepackage{graphicx}
\definecolor{light}{rgb}{0.5, 0.5, 0.5}
\def\light#1{{\color{light}#1}}
\usepackage{multibib}
\usepackage{hyperref}
\usepackage{filecontents}
\pgfplotsset{compat=newest}

\newcommand*{\MinNumber}{-5}%
\newcommand*{\MidNumber}{0}%
\newcommand*{\MaxNumber}{30}%
\newcommand{\ApplyGradient}[1]{%
        \ifdim #1 pt > \MidNumber pt
            \pgfmathsetmacro{\PercentColor}{max(min(100.0*(#1 - \MidNumber)/(\MaxNumber-\MidNumber),100.0),0.00)} %
            \hspace{-0.33em}\colorbox{red!\PercentColor!white}{#1}
        \else
            \pgfmathsetmacro{\PercentColor}{max(min(100.0*(\MidNumber - #1)/(\MidNumber-\MinNumber),100.0),0.00)} %
            \hspace{-0.33em}\colorbox{green!\PercentColor!white}{#1}
        \fi
}

\newcommand{\DEVELOPMENT}{1} 
\usepackage{ifthen}
\ifthenelse{\DEVELOPMENT = 1}{
	\newcommand{\pa}[1]{\textcolor{blue}{\textbf{PA:} #1}}		
	\newcommand{\md}[1]{\textcolor{red}{\textbf{MD:} #1}}
 	\newcommand{\tz}[1]{\textcolor{cyan}{\textbf{TZ:} #1}}
 	\newcommand{\am}[1]{\textcolor{purple}{\textbf{AM:} #1}}
 	\newcommand{\bi}[1]{\textcolor{green}{\textbf{BM:} #1}}
 	\newcommand{\ps}[1]{\textcolor{orange}{\textbf{PS:} #1}}
 	
}{
    \newcommand{\bi}[1]{}
	\newcommand{\md}[1]{}		
	\newcommand{\pa}[1]{}		
	\newcommand{\tz}[1]{}
	\newcommand{\am}[1]{}
	\newcommand{\ps}[1]{}
}

\AtBeginDocument{%
  \providecommand\BibTeX{{%
    \normalfont B\kern-0.5em{\scshape i\kern-0.25em b}\kern-0.8em\TeX}}}

\copyrightyear{2023}
\acmYear{2023}
\setcopyright{acmlicensed}
\acmConference[WWW '23]{Proceedings of the ACM Web Conference 2023}{May 1--5, 2023}{Austin, TX, USA}
\acmBooktitle{Proceedings of the ACM Web Conference 2023 (WWW '23), May 1--5, 2023, Austin, TX, USA}
\acmPrice{15.00}
\acmDOI{10.1145/3543507.3583356}
\acmISBN{978-1-4503-9416-1/23/04}

\begin{document}

\title{HateProof: Are Hateful Meme Detection Systems really Robust?}

\author{Piush Aggarwal}
\authornote{Equal Contribution.}
\email{piush.aggarwal@fernuni-hagen.de}
\affiliation{%
  \institution{CATALPA, FernUniversit{\"a}t in Hagen}
  \city{Hagen}
  \country{Germany}}

\author{Pranit Chawla}
\authornotemark[1]
\email{pranitchawla98@iitkgp.ac.in}
\orcid{1234-5678-9012}
\author{Mithun Das}
\authornotemark[1]
\email{mithundas@iitkgp.ac.in}
\affiliation{%
  \institution{Indian Institute of Technology Kharagpur}
  \city{Kharagpur}
  \state{West Bengal}
  \country{India}
  \postcode{721302}
}

\author{Punyajoy Saha}
\email{punyajoys@iitkgp.ac.in}
\affiliation{%
  \institution{Indian Institute of Technology Kharagpur}
  \city{Kharagpur}
  \state{West Bengal}
  \country{India}
  \postcode{721302}
}

\author{Binny Mathew}
\email{binnymathew@iitkgp.ac.in}
\affiliation{%
  \institution{Indian Institute of Technology Kharagpur}
  \city{Kharagpur}
  \state{West Bengal}
  \country{India}
  \postcode{721302}
}

\author{Torsten Zesch}
\email{torsten.zesch@fernuni-hagen.de}
\affiliation{%
  \institution{CATALPA, FernUniversit{\"a}t in Hagen}
  \city{Hagen}
  \country{Germany}}

\author{Animesh Mukherjee}
\email{animeshm@cse.iitkgp.ac.in}
\affiliation{%
  \institution{Indian Institute of Technology Kharagpur}
  \city{Kharagpur}
  \state{West Bengal}
  \country{India}
  \postcode{721302}
}

\renewcommand{\shortauthors}{Aggarwal, et al.}

\begin{abstract}
Exploiting social media to spread hate has tremendously increased over the years. Lately, multi-modal hateful content such as memes has drawn relatively more traction than uni-modal content. Moreover, the availability of implicit content payloads makes them fairly challenging to be detected by existing hateful meme detection systems. In this paper, we present a use case study to analyze such systems' vulnerabilities against external adversarial attacks. We find that even very simple perturbations in uni-modal and multi-modal settings performed by humans with little knowledge about the model can make the existing detection models highly vulnerable. Empirically, we find a noticeable performance drop of as high as 10\% in the macro-F1 score for certain attacks. As a remedy, we attempt to boost the model's robustness using contrastive learning as well as an adversarial training-based method - \emph{VILLA}. Using an ensemble of the above two approaches, in two of our high resolution datasets, we are able to (re)gain back the performance to a large extent for certain attacks. We believe that ours is a first step toward addressing this crucial problem in an adversarial setting and would inspire more such investigations in the future.
\end{abstract}




\maketitle


\section{Introduction}



Leveraging the doctrine of freedom of speech~\citep{10.1145/3184558.3191531,doi:10.1080/13600834.2018.1494417,Guiora2017}, misanthropists are spreading hatred in the society using social media platforms. Memes have been widely used in the online world to spread harmful content at an alarming rate. To detect such malicious content, social media companies employ moderators to manually screen posts publicized on their platforms. However, due to the high volume of content dissemination, it has become challenging to label such contents manually. Hence, automatic moderation techniques are required. With the advancements in NLP and vision technologies machines are now able to interpret the semantics of hate, thereby, facilitating deceleration of the spread of such content~\citep{sabat2019hate, aggarwal-etal-2021-vl, https://doi.org/10.48550/arxiv.2012.07788, Zhu2020FreeLB:, https://doi.org/10.48550/arxiv.2012.12871, https://doi.org/10.48550/arxiv.2012.12975}.


\begin{figure}
  \begin{center}
  \resizebox{0.35\textwidth}{!}{%
\includegraphics[]{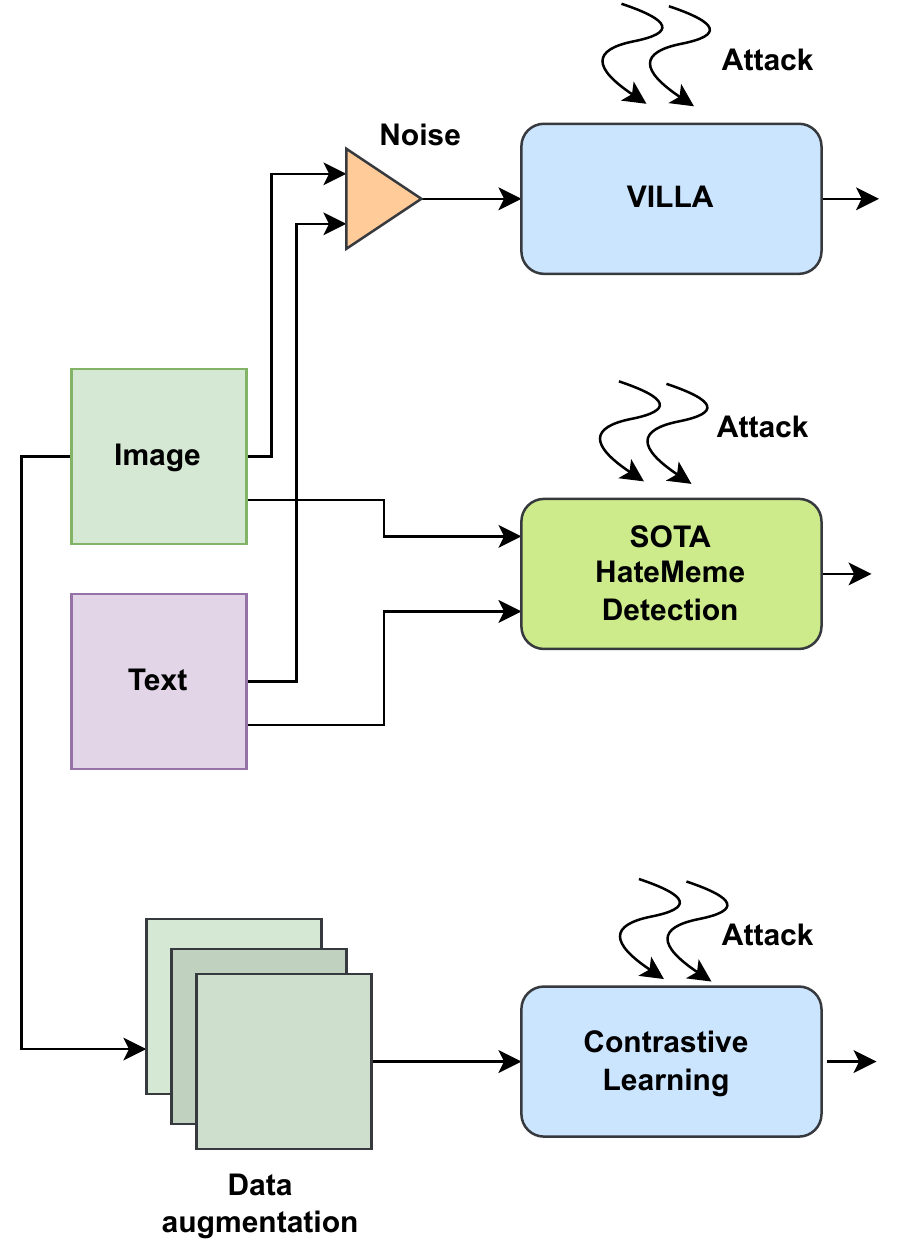}
}
        \end{center}
        \caption{A schematic showing the overall setup of our experiments.}

        \label{schematic}
\end{figure}
However, systems like these that rely more on machine learning than empirical principles are oftentimes susceptible to unexpected outcomes. A critical difference between multimodal systems and their unimodal counterparts is the fusion mechanism \cite{https://doi.org/10.48550/arxiv.2112.12792}. This fusion mechanism merges multiple input modalities to learn their common representation, which is then processed by numerous fully linked layers (assuming any popular deep learning setup) to predict classification results depending on the nature of the corresponding downstream tasks. The system adopts multiple strategies to learn strong fusion embeddings of the input modalities. This fusion mechanism presents a new challenge for studying the robustness of these models to adversity. Hateful memes comprise implicit properties, which any of their modalities can acquire. As a result, the possibility of experiencing unexpected events is more. Consequently, performance could be potentially localized and the model may get biased \cite{https://doi.org/10.48550/arxiv.2204.01734} thus posing significant challenges. 

Misuse of hate meme detection systems is a common problem similar to deliberate misconduct in AI-based biometric authentication systems \citep{toygar2020fyo, wang2019multi}. Evtimov et. al. \cite{https://doi.org/10.48550/arxiv.2011.12902} compares the model vulnerabilities toward attacks that are attempted when full knowledge of the model is available to the one having partial or no access to model properties. In this paper, we compose adversarial attacks based on partial knowledge of the model \cite{https://doi.org/10.48550/arxiv.2011.12902} where attackers are assumed to know that the model takes clues from text tokens and image pixels to learn the hate classifier, 
and propose nine different attacks that can be easily integrated with any existing meme while maintaining the message perception. The attacks are applied to the individual as well as both of the modalities of the meme. One of the key aspects of our analysis is to consider hateful meme detection as an end-to-end task. Therefore, we employ OCR technology to extract meme texts rather than directly using the text provided in the datasets. We find that the state-of-the-art models are highly vulnerable to the attacks designed by us. We, therefore, propose \emph{adversarial training} and \emph{contrastive learning} based countermeasures to tackle such vulnerabilities. Figure~\ref{schematic} illustrates our overall experimental pipeline.



The major contributions of this paper include the following.

\begin{compactitem}
    \item We extend the study in \cite{https://doi.org/10.48550/arxiv.2011.12902} to examine the model's vulnerability to partial model knowledge-based adversarial attacks. In particular, we propose nine different attacks to investigate the vulnerabilities of the existing hate meme detection systems to human-induced adversarial attacks. We find that all of these models are highly vulnerable with a drop of as high as 10\% in macro-F1 performance in certain cases.
    \item We develop two different countermeasures to tackle such adversarial inputs. Precisely, we implement a variant of \textsc{SimCLR} \cite{chen2020big} which is based on contrastive learning as well as the popular \textit{VILLA} \cite{NEURIPS2020_49562478} approach based on adversarial training. In the former case, a function learns additional signals from simple augmentations of each data point in the training set and that is blended with hate meme detection model's objectives through a suitable loss function thus increasing the overall generalization capabilities of the model. In the latter case, in addition to available data points, a model is trained on adversarial examples that are generated by perturbing one modality of the input data while keeping another one unchanged at a time. For image modality, unlike \cite{https://doi.org/10.48550/arxiv.1312.6199} where pixels are perturbed, we consider image-region features. Similarly, the final embedding layer is used for textual tokens, where a minute amount of noise is introduced. 
    \item In addition, we also propose an \textsc{Ensemble} of the above two approaches to salvage the best of both of them. We observe that for two of our high resolution datasets, the \textsc{Ensemble} based countermeasure is able to successfully tackle various different forms of adversarial attacks. The \textit{VILLA} approach is a close second after the \textsc{Ensemble} approach.
    \end{compactitem}
    
\noindent
In summary, our analysis\footnote{\url{https://github.com/aggarwalpiush/Robust-hatememe-detection}} reveals the plausible loopholes that potential haters can adopt in order to escape from the detection systems. Given this, we play proactively and propose generalized countermeasures that are efficient under many attack conditions.






\section{Related work}

\noindent\textbf{Hate meme detection}: Detecting hate memes has become an increasingly popular research topic owing to the alarming growth in hateful memes across different social media platforms. The models often perform multimodal pre-training on huge unsupervised corpus \cite{https://doi.org/10.48550/arxiv.2009.08395} followed by a fine-tuning on a relatively smaller set of supervised hate data. In the case of memes, the early and late fusion of features belonging to each modality is done before generating the final predictions. 
In the Hateful Memes challenge \cite{https://doi.org/10.48550/arxiv.2005.04790} most effective systems featuring at the top of the leaderboard were heavily relying on large multimodal transformer models such as \textsc{VL-BERT} \citep{Zhu2020FreeLB:, aggarwal-etal-2021-vl}, \textsc{Oscar} \cite{li2020oscar}, \textsc{Uniter} \cite{https://doi.org/10.48550/arxiv.2012.12871}, \textsc{VisualBERT} \cite{https://doi.org/10.48550/arxiv.2012.12975}, and \textsc{LXMERT} \cite{https://doi.org/10.48550/arxiv.1908.07490}. Recently many works have also proposed an ensemble of different visual-linguistic models for obtaining better detection performance \citep{Zhu2020FreeLB:, https://doi.org/10.48550/arxiv.2012.07788, aggarwal-etal-2021-vl}. 


\noindent\textbf{Datasets}: Multiple datasets \citep{https://doi.org/10.48550/arxiv.2005.04790, pramanick-etal-2021-detecting,fersini2022semeval, suryawanshi-etal-2020-multimodal} have been built to bring in more diversity in the detection task. In addition, there are other similar datasets that emphasize sentiment, humor, offensiveness, and motive of memes \cite{sharma-etal-2020-semeval}. Apart from English, memes have also been analyzed in other languages datasets \citep{Miliani2020,gold_aggarwal_zesch_2021}.

\noindent\textbf{Adversarial attacks}: Previous studies raise a concern about model biasing and domain-specific responses \cite{https://doi.org/10.48550/arxiv.2204.01734}. Among images, tampering has been used for the malicious purpose for quite sometime now. Bayar et.al \citep{10.1145/2909827.2930786} embeded noise using image processing functions namely scaling, blurring and introduction of white noise. For automatic image alteration, GAN-based systems are commonly used \cite{8397040}. Color filter array (CFA) based spectroscopy has been performed to detect pixel level abnormalities \cite{GonzlezFernndez2018}. Further, several perturbation-based algorithms have been proposed to attack image modality~\cite{7958570}. Among text-based attacks, \cite{https://doi.org/10.48550/arxiv.1808.09115} added \emph{LOVE} to every text input and found that the nature of the data and the labeling standards are more crucial than the model architecture. The study showed that character-level feature training helped to uplift model robustness. Content code blurring has also been used to both emphasize \cite{4624687} and forge \cite{9540787} the textual quality. Attacks also have been developed while considering both modalities. For example, \cite{vishwamitra2021understanding} observed that decoupling of image and text components directly defeats the sole purpose of the visual-linguistic models.

In this paper, unlike \cite{https://doi.org/10.48550/arxiv.2011.12902}, we specifically focus on the model-independent attacks which humans can introduce with little technical expertise\footnote{They are aware of using internet facilities.}. Attack for instance adding noise to the image and texts in the memes can be easily implemented without prior knowledge of the classification models. Unlike previous work which provides a very abstract understanding of this noise category, we attempt to broaden the attack's horizon at a fine-grained level and analyze their impact on a variety of hate meme detection model paradigms.  

\noindent\textbf{Multi-modal robustness}
Earlier work in model robustness were generally limited to uni-modal settings with end-to-end adversarial training (AT) \citep{https://doi.org/10.48550/arxiv.1705.07204, https://doi.org/10.48550/arxiv.1706.06083, Xie_2019_CVPR, NEURIPS2020_5898d809, pmlr-v97-zhang19p, https://doi.org/10.48550/arxiv.1808.09115} where training data was augmented with synthetic data to facilitate provably robust training. 
For multi-modal cases, Yang et. al. \cite{Yang_2021_CVPR} emphasized the robust fusion of modalities rather than end-to-end parameter training. Recent studies show that, by injecting adversarial perturbations into feature space, AT can further improve model generalization on language understanding \citep{Zhu2020FreeLB:, jiang-etal-2020-smart} as well as visual-linguistic tasks \cite{NEURIPS2020_49562478}. Exploiting contrastive learning to improvise model robustness exhibits impressive performance in text classification tasks \citep{qiu-etal-2021-improving, https://doi.org/10.48550/arxiv.2107.10137}. In addition, there are approaches \citep{li2021closer, gan2020largescale, NEURIPS2020_49562478} which insert visual-linguistic perturbation into the embedding space of the models to increase the adversarial examples in the input. In this work, we intend to exploit contrastive learning proposed by Chen et.al. \citep{chen2020simple} to augment the robustness of hate meme classification. 
In another countermeasure that we develop, unlike \cite{NEURIPS2020_49562478}, which introduces uniformly distributed perturbation in the embedding space of both modalities, we use Gaussian noise based perturbations to generate the adversarial examples to perform the training. 

\section{Datasets}
In order to  analyze the vulnerability of available hate meme classifiers as well as evaluate our countermeasure proposal, we have used three benchmark datasets (see Table~\ref{dataset}). Note that we apply the proposed adversarial attacks only on the test set in order to analyze the model robustness.
\begin{compactitem}
    \item Kiela et.al.  \cite{https://doi.org/10.48550/arxiv.2005.04790} (\textsc{Fbhm}) contains 10,000 memes collected from Getty images and are semi-artificially annotated using benign confounders. The dataset consists of five varieties of memes. These include \emph{multimodal hate} where both modalities possess benign confounders, \emph{unimodal hate} where at least one of the modalities is already hateful, \emph{benign image} as well as \emph{benign text} confounders and \emph{random not-hateful} examples. First four categories are annotated with \emph{hateful} label and rest with \emph{non hateful} label. Since the meme text is also annotated here, no error is induced during the OCR process. The dataset is split into 85\% training, 5\% development, and 10\% test sets. The development and test set are fully balanced with fixed proportions of each variety discussed above. 
    \item Pramanick et.al. \cite{pramanick-etal-2021-detecting} (\textsc{HarMeme}) contains COVID-related memes posted in US social media. Some of the query keywords include \emph{Wuhan virus}, \emph{US election}, \emph{COVID vaccine}, \emph{work from home} and \emph{Trump not wearing mask}. Unlike \cite{https://doi.org/10.48550/arxiv.2005.04790}, all of the memes are original and shared across social media. Also, the associated textual content is the output from Google Vision API rather than the manual. The resolution from the memes is also preserved. We group the whole dataset into two categories -- (i) \emph{hateful} which consists of both \emph{harmful} and \emph{partially harmful} labelled memes provided in the dataset, and (ii) \emph{non hateful} which consists of memes tagged as \emph{non harmful} in the dataset. Out of a total of 3,544 data points, we again use 85\%, 5\%, and 10\% for training, validation and test respectively.
    \item Fersini et. al. \cite{fersini2022semeval} (\textsc{Mami}) We focus only on SUBTASK-A of the dataset where memes are either labelled as \emph{misogynist} or \emph{non misogynist}. We relabel the former as \emph{hateful} and the latter as \emph{non-hateful} making it consistent with our setup. The memes have been collected from social media websites and pertain to threads with women as subjects, or having antifeminist content. Such memes having famous women personalities such as Scarlett Johansson, Emilia Clarke, etc. as well as hashtags such as \texttt{\#girl}, \texttt{\#girlfriend}, \texttt{\#women}, \texttt{\#feminist} are also considered. Likewise \cite{pramanick-etal-2021-detecting}, meme text is extracted with Google Vision API. In total, we use a balanced set of 10,000 instances. Out of this, we use 10\% posts each for development and test which are randomly stratified. 
\end{compactitem}

\begin{table}[t]
\resizebox{\linewidth}{!}{%
\begin{tabular}{@{}lllll@{}}
\toprule
  & \textsc{Fbhm}~\cite{https://doi.org/10.48550/arxiv.2005.04790} & \textsc{HarMeme}~\cite{pramanick-etal-2021-detecting} & \textsc{Mami}~\cite{fersini2022semeval} \\ 
 \cline{2-4}
Train/Dev/Test  & 8500/500/1000   & 3013/177/354  & 8000/1000/1000 \\
Hate \% & 37.56   & 26.21  & 50 \\
Domain & \emph{in the wild}  & Covid-19/US Election & Misogynistic \\ 
Bit depth (Avg.) & 9.54 & 43.90 & 4.30 \\ \bottomrule
\end{tabular}
}
\caption{Dataset statistics.}
\label{dataset}
\end{table}

\section{Adversarial Attacks}

\begin{figure*}
     \begin{subfigure}[c]{.36\linewidth}
         \centering
         \includegraphics[width=\linewidth]{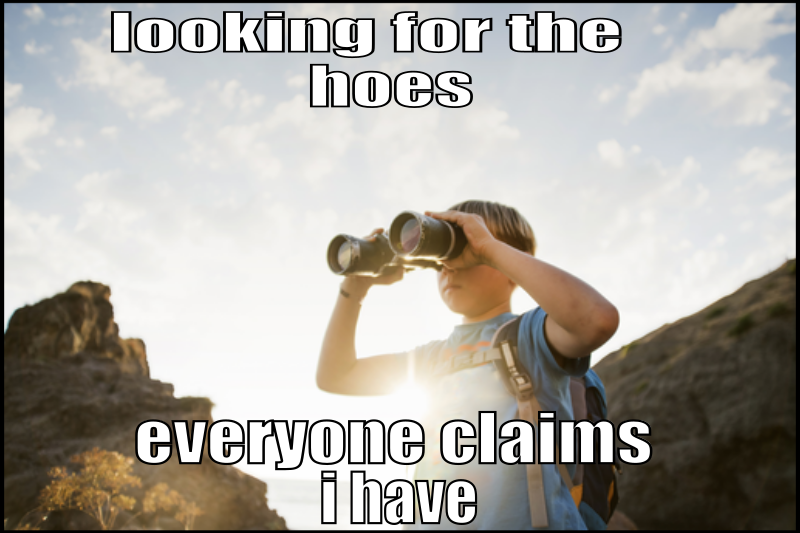}
         \caption{Original meme.}
         \label{orig}
     \end{subfigure}
     \hfill
     \begin{tabular}[c]{@{}c@{}}
     \begin{subfigure}[c]{.20\linewidth}
         \centering
         \includegraphics[width=\linewidth]{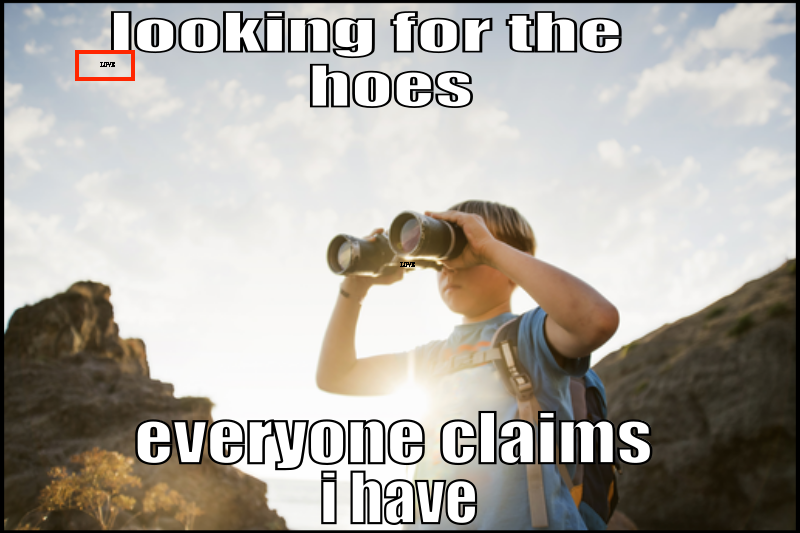}
         \caption{Text: \texttt{Add}.}
         \label{textadd}
     \end{subfigure}
\hfill
    \begin{subfigure}[c]{.20\linewidth}
         \centering
         \includegraphics[width=\linewidth]{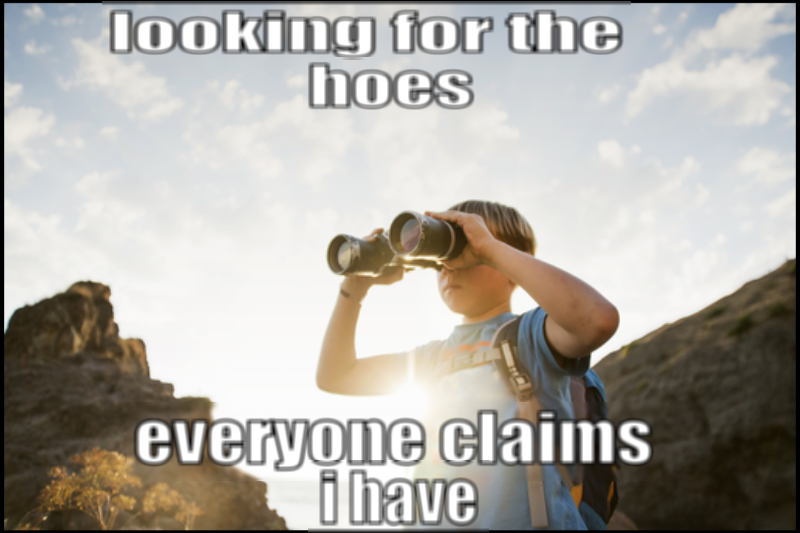}
          \caption{Text: \texttt{Blur}.}
         \label{textblurr}
     \end{subfigure}
\hfill
    \begin{subfigure}[c]{.20\linewidth}
         \centering
         \includegraphics[width=\linewidth]{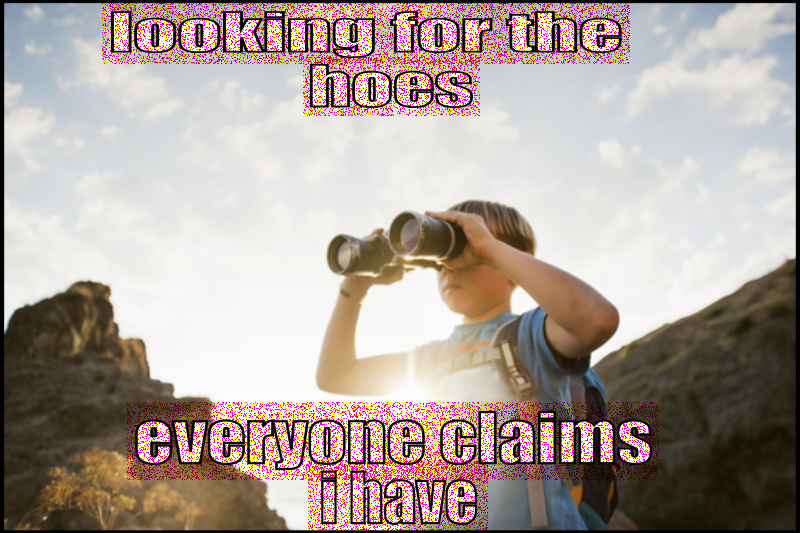}
          \caption{Text: \texttt{SaltPepper-T}.}
         \label{textsp}
     \end{subfigure}\\
\noalign{\bigskip}%
    \begin{subfigure}[c]{.20\linewidth}
         \centering
         \includegraphics[width=\linewidth,page=2]{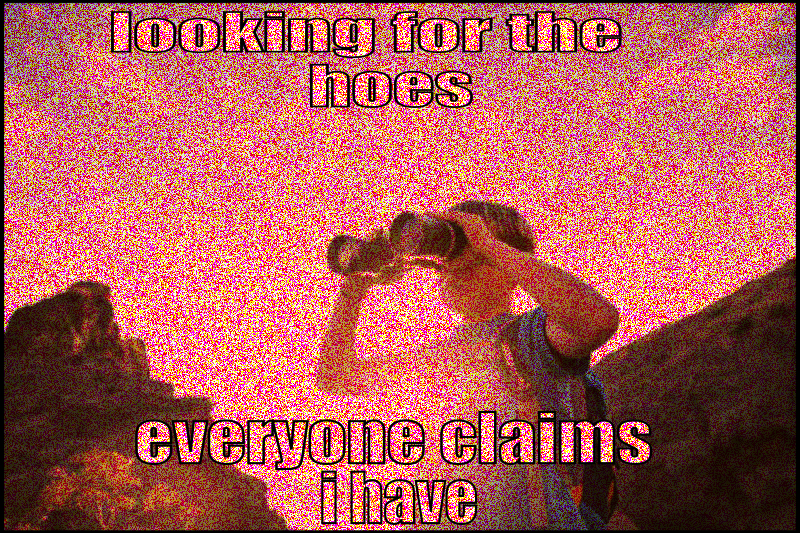}
          \caption{Image: \texttt{SaltPepper-I}.}
         \label{imgsp}
     \end{subfigure}

    \hfill
    \begin{subfigure}[c]{.20\linewidth}
         \centering
         \includegraphics[width=\linewidth]{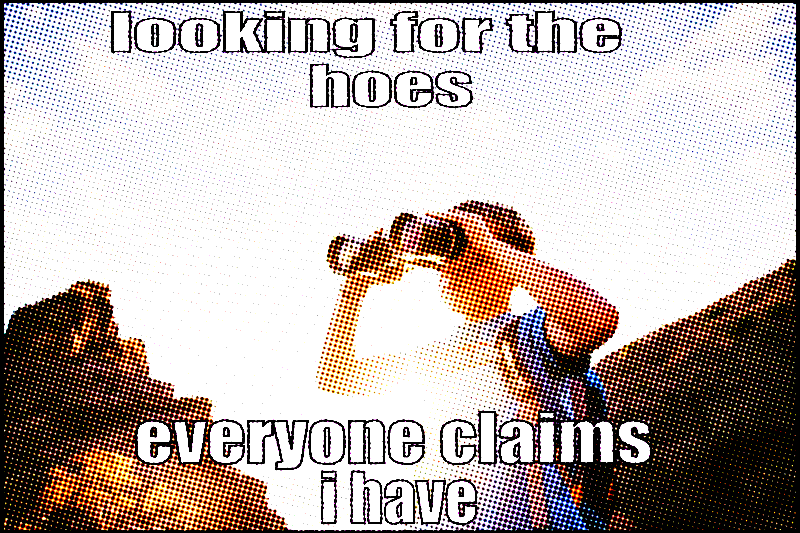}
          \caption{Image: \texttt{Newsprint}.}
         \label{textsp}
     \end{subfigure}
     \hfill
    \begin{subfigure}[c]{.20\linewidth}
         \centering
         \includegraphics[width=\linewidth]{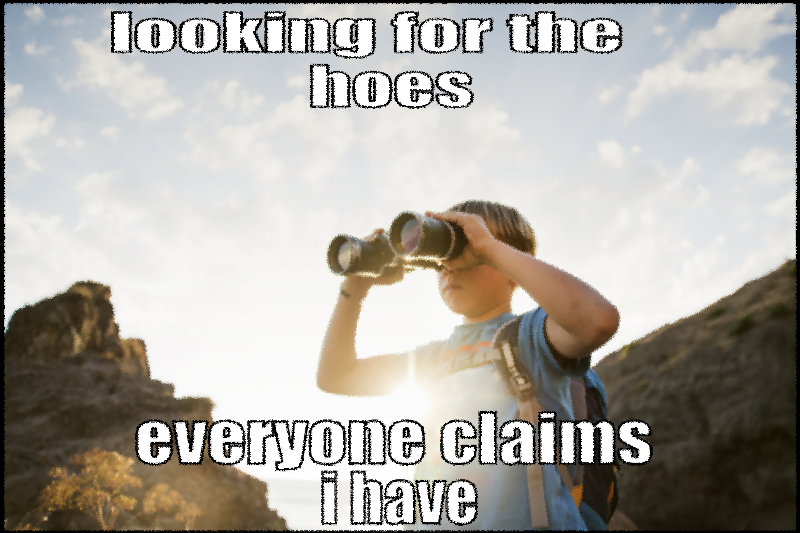}
          \caption{Image: \texttt{Spread}.}
         \label{textsp}
     \end{subfigure}\\
        
     \end{tabular}
        \caption{Different forms of adversarial attacks that could presented to a hate meme detection model.}

        \label{adversrialattacks}
\end{figure*}

Szegedy et.al \cite{https://doi.org/10.48550/arxiv.1312.6199} pointed out that neural networks are highly susceptible to minor changes in the input aka adversarial attacks unless they are properly trained to handle such changes. The human eye is almost insensitive to these minor changes; however, they are capable of hindering the prediction probabilities of the neural classifiers. Adversarial attacks can be \emph{whitebox} where internal parameter and gradients of the models are known. However, in the case of \emph{blackbox} attacks, the model is considered as an oracle and attacks are developed based on model confidence and output prediction probabilities. Therefore, in this case, attacks can be implemented in unlimited settings. As corporate models are not disclosed and updated time to time, therefore attacks in even blackbox settings can vary. In this work, we consider attacks based on partial knowledge of the models. Therefore, we compose model independent attacks and categorize them in nine different types including simple attacks such as \texttt{SaltPepper} noise which has shown noticeable effects in the past \citep{Jaiswal_Duggirala_Dash_Mukherjee_2022, Dooley2021RobustnessDI}. Figure~\ref{adversrialattacks} illustrates example of attacks chosen for our study.

\subsection{Attacks on text} Unlike previous work \cite{https://doi.org/10.48550/arxiv.2011.12902} where edit distance-based attacks were chosen, in our case, we apply textual attacks in such a way that directly affects the recognition quality of OCRs applied to the memes to extract text. Such attacks study the importance of text modality in real-time settings. The different attack forms include

\begin{compactitem}
    \item \texttt{Add}: Adding high polarity token to the input text often misleads the classifier \cite{https://doi.org/10.48550/arxiv.1808.09115}. Taking cue from this paper, we consider the token \emph{LOVE} to have a small font of size five pixels and embed it at random locations of the input image. To establish the generalizability we also repeat the attack for a (semantically) related token \emph{CARESS}.

    \item \texttt{Blur}: In addition to visual-linguistic models, OCRs in the pre-processing stage also play important role in the overall performance of the hateful meme detection problem. Diminishing the quality of the textual part can hamper the recognition quality of the OCR models. We blur the text using the openCV tool\footnote{\url{https://github.com/opencv/opencv-python}}.
    \item \texttt{SaltPepper-T}: Likewise blurring, we add salt-and-pepper noise to only the text area of the image. To do so, we extract the bounding box where text is available using easyOCR\footnote{\url{https://github.com/JaidedAI/EasyOCR}} python library. Then we randomly add white and black pixels only to the area contained in this box.
\end{compactitem}

\subsection{Attacks on image} Cases where the internal behavior of the threat model is not predetermined, the attacker or the adversary cannot use any gradient information to generate appropriate attacks. However, they can introduce random modifications to the image while preserving its message. Therefore we apply image distortions using the \textsc{Gimp} software\footnote{\url{https://www.gimp.org/}} in different settings.

\begin{compactitem}
    \item \texttt{SaltPepper-I}: White and black pixels are injected into the whole image in such a way that neither the image perception nor the message conveyed is compromised. We inject the pixels in two different settings (\emph{high} and \emph{low}).
    \item \texttt{Newsprint}: In this case, the image is poisoned with clustered-dot dithering to get a feel of newspaper printing.
    \item \texttt{Spread}: Swapping of each pixel with another randomly chosen pixel is done in order to get a slightly jittery output. Here again, both \emph{high} and \emph{low} amounts of spreading have been done.  
\end{compactitem}

\noindent
In addition, we consider combining text as well as image-based attacks. To do so, we choose the common attack scenarios in each modality and aggregate \texttt{SaltPaper-I} (commonly used in photographs \cite{lendave_2021}) and \texttt{Add} (token \emph{LOVE} in our case \cite{https://doi.org/10.48550/arxiv.1808.09115}) in order to analyze the model vulnerability against joint modality attacks.

\section{Countermeasures}

In this section, we discuss two different countermeasures that we employ to tackle such adversarial attacks one of which is based on contrastive learning while the other on adversarial training. We keep the model agnostic of the attack type since otherwise it would learn the properties of certain types only and would lose generalizability. All our attack types are thus unseen to the model. 

\begin{figure*}
     \begin{subfigure}[c]{.19\linewidth}
         \centering
         \includegraphics[width=\linewidth]{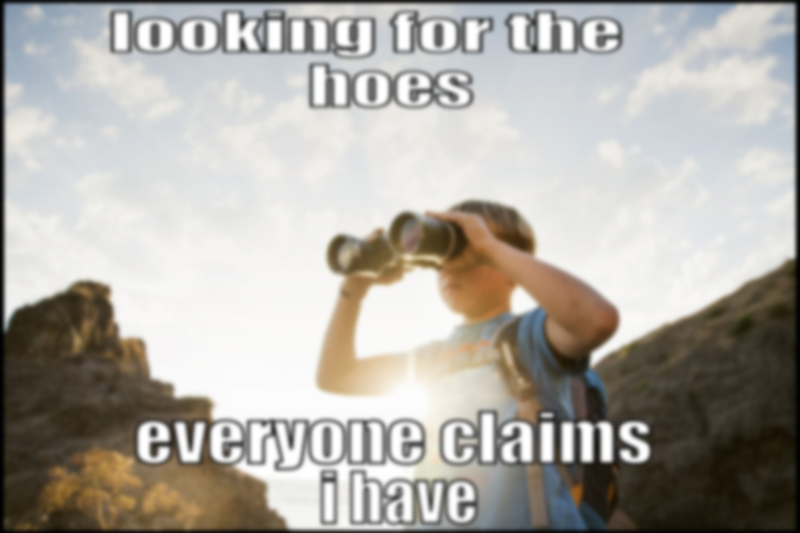}
         \caption{Blurring.}
         \label{orig}
     \end{subfigure}
     \hfill
     \begin{subfigure}[c]{.19\linewidth}
         \centering
         \includegraphics[width=\linewidth]{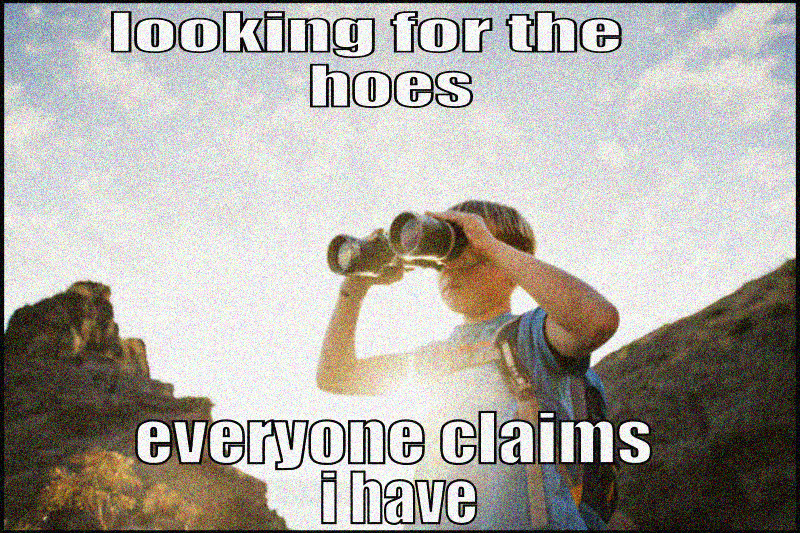}
         \caption{Random noising.}
         \label{textadd}
     \end{subfigure}
\hfill
    \begin{subfigure}[c]{.19\linewidth}
         \centering
         \includegraphics[width=\linewidth]{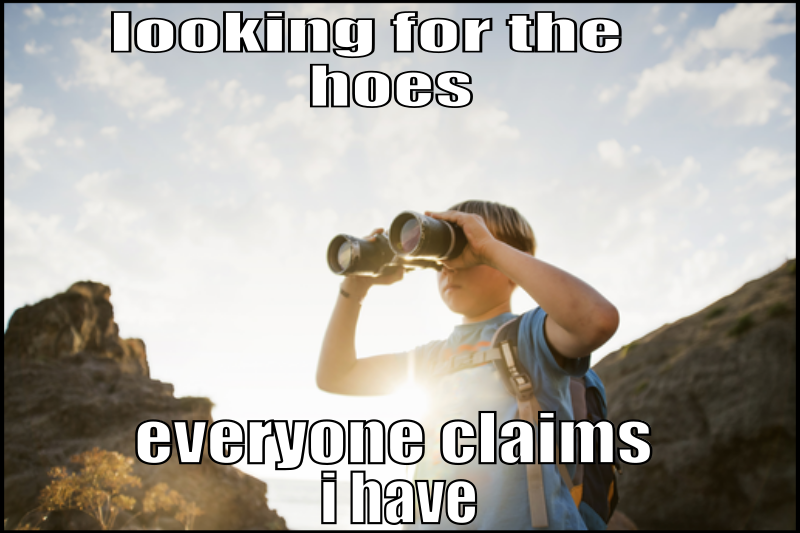}
          \caption{Colour jittering.}
         \label{textblurr}
     \end{subfigure}
\hfill
    \begin{subfigure}[c]{.19\linewidth}
         \centering
         \includegraphics[width=\linewidth]{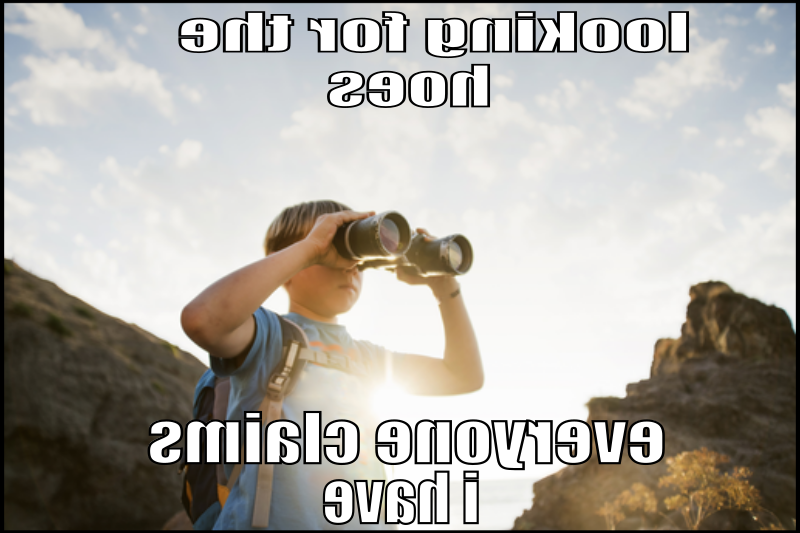}
          \caption{Horizontal flipping.}
         \label{textsp}
     \end{subfigure}
\hfill
    \begin{subfigure}[c]{.19\linewidth}
         \centering
         \includegraphics[width=\linewidth]{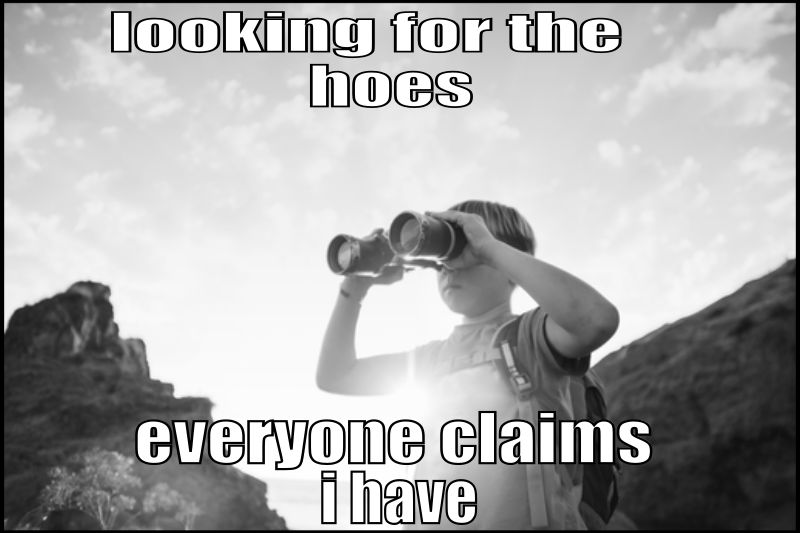}
          \caption{Grey scaling.}
         \label{textsp}
     \end{subfigure}     
\\
    \begin{subfigure}[c]{.25\linewidth}
         \centering
         \includegraphics[width=\linewidth]{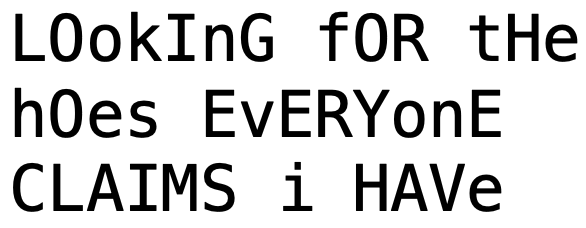}
          \caption{Casing alteration.}
         \label{imgsp}
     \end{subfigure}
    \hfill
    \begin{subfigure}[c]{.25\linewidth}
         \centering
         \includegraphics[width=\linewidth]{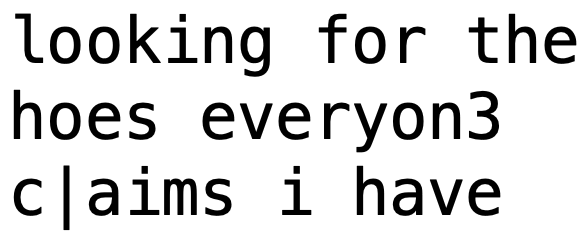}
          \caption{Similar character replacement.}
         \label{textsp}
     \end{subfigure}
     \hfill
    \begin{subfigure}[c]{.25\linewidth}
         \centering
         \includegraphics[width=\linewidth]{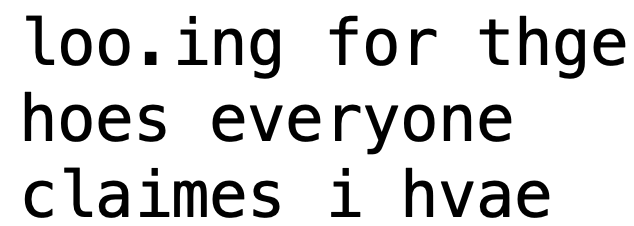}
          \caption{Typo induction.}
         \label{textsp}
     \end{subfigure}\\
        
        \caption{Representative examples of image and text variations used for synthetic augmentation in contrastive learning.}

        \label{contlearning}
\end{figure*}

\subsection{Contrastive learning through data augmentation} 
\label{dataaug}
\emph{Data augmentation} \citep{https://doi.org/10.48550/arxiv.2111.05328, https://doi.org/10.48550/arxiv.2103.01946} is the one among those promising methods that are prominently used to improve model generalization and robustness. The main idea is to introduce a small mutation in the original training data and synthetically generate new samples to virtually increase the amount of training data \cite{Simard2012}. The points are generally represented as interpolated output of actual training samples. This resolve under-determined and poor generalization problems that occur in the model when they are trained on scarce dataset \cite{8545506}. During the formalization of augmented samples, human knowledge is needed to describe a neighborhood around each instance in the training data \cite{NIPS2000_ba9a56ce}. For instance, during the image classification process, a variety of image transformation operations are performed such as horizontal reflections, slight rotations, and mild scalings. However, sharing the same class by each neighbor examples restrict the models to learn the vicinity relationships across different classes \cite{8545506}. Therefore, in this work, we consider contrastive learning (CL), which represents a function learned on simple augmentations of each data point in the training set guided via a suitable loss function. The loss function is derived in such a way that it maximizes the agreement among the augmentations of a single data point while spreading apart the augmentations from different data points. The loss function can later be integrated into any classifier to enhance its generalization capabilities \cite{chen2020simple}. In this work, we apply the \emph{SimCLR} \cite{chen2020big} implementation of contrastive learning with crucial revisions to adapt it to the multi-modal settings (memes in our case). 


\begin{algorithm}
\caption{Computation of the contrastive loss function $\mathcal{L}$}\label{alg:cap}
\begin{flushleft}
\textbf{Input:} \texttt{batch size $N$, temperature constant $\tau$, pretrained encoder network $f$, projection generator $g$, Augly function $A$, instance $x$} \\
\end{flushleft}
\begin{algorithmic}[1]
\ForAll{\texttt{$i$ $\epsilon$ {1, . . . , N}}}
\State \light{\# Learn representation}
\State \texttt{$z_i = g(f(x_{i}))$}
\State apply Augly \texttt{$x'_{i} \leftarrow A(x_{i})$}
\State \texttt{$h_{j} = f(x'_{i})$}
\State \texttt{$z_{j} = g(h_{j})$}
\State \light{\# Compute pairwise cosine similarity}
\State \texttt{$s_{(i,j)}$ = $z_{i}^Tz_{j}\slash(||z_{i}||||z_{j}||)$ }
\State \texttt{$s_{(j,i)}$ = $z_{j}^Tz_{i}\slash(||z_{i}||||z_{j}||)$ }

\State \light{\# Define loss function $\ell(i,j)$}
\State \texttt{$\ell(i,j)$ = $- log\frac{exp(s_{(i,j)}\slash\tau)}{\sum_{n=1}^{N} \mathbbm{1}_{[k \neq i]} exp(s_{(i,k)}\slash\tau)}$}
\EndFor
\State \light{\# Constrastive loss function $\mathcal{L}$}
\begin{flushleft}
\Return  
\texttt{$\mathcal{L}$ = $\frac{1}{2N}\sum_{k=1}^{N}[\ell(2k-1,2k) + \ell(2k, 2k-1)]$}  
\end{flushleft}
\end{algorithmic}
\label{contloss}
\end{algorithm}


To implement CL, the first step would be to obtain synthetic augmentations. Therefore we use the \emph{Augly} \cite{papakipos2022augly} python library to obtain such augmentations. In case of the image component, for each input sample we choose one of the noisy versions among \texttt{blurring}, \texttt{random noising}, \texttt{color jittering}, \texttt{horizontal flipping} or \texttt{gray scaling}. Figure~\ref{contlearning} illustrates the examples of augmented samples. Similarly to generate textual samples, we use \emph{Augly text}. Three different types of text augmentation that we use are \texttt{casing alteration}, \texttt{similar character replacement} as well as \texttt{typo induction}. In the end, we randomly combine augmented versions from different modalities to produce the related sample for each input instance. We redefine the loss function of the (best performing) state-of-the-art hate meme classification model to incorporate an additional contrastive loss estimated from the augmented data. Precisely, in addition to the cross entropy loss used during the training of the classification model, we add the contrastive loss function computed in Algorithm~\ref{contloss}. We calculate the encoder network's projection $g$ (aka the embedding) for each input instance $x$ and its augmented version $A(x)$. Next we calculate the cosine similarity between the projections. Subsequently, contrastive loss function is calculated as shown in line 11 of the algorithm where the numerator encapsulates the similarity of a given instance (aka positive example) and the denominator encapsulates the similarity between pairs of remaining instances of the batch which act as negative examples. $\mathbbm{1}_{[k \neq i]}$ is an indicator function which is $1$ for negative examples and $\tau$ denotes the temperature parameter. 

\subsection{Adversarial training with VILLA}
\label{advtraining}
Adversarial training has shown a promising role to counter adversarial attacks in machine learning models. Earlier analysis of model robustness has been limited to performance on surprising altercations appearing in relatively clean data~\cite{10.1007/978-3-642-40994-3_25, https://doi.org/10.48550/arxiv.1312.6199}. However, robustness against deliberate obfuscation activities, for instance human induced adversarial attacks on hate meme, has recently emerged as a critical problem. A robust classifier should be able to accurately identify adversarially  poisoned images. Training the model with augmented samples as discussed in the previous section can be helpful in order to surpass the vulnerabilities of the existing hate meme detection models. However, due to limited flexibility in explicit augmented example generation, the robustness can only be increased up to a certain extent. Therefore, here, we attempt to generate adversarial examples by performing perturbation in the embedding space \cite{Zhu2020FreeLB:} of both the image and the text for each training input instance. For textual and image parts, we use BERT and regional embeddings respectively \cite{10.5555/3495724.3496279}. To do so, we implement the \emph{VILLA} framework \cite{NEURIPS2020_49562478}\footnote{\url{https://github.com/zhegan27/VILLA}} where one modality is perturbed keeping the other one unchanged at a time. For image modality, unlike \cite{https://doi.org/10.48550/arxiv.1312.6199} where pixels are perturbed, we consider image-region features. Similarly for textual tokens, final embedding layer is used where a small amount of noise is introduced.
 Small perturbations in the feature space keep the instances within the classification boundary and act as a form of latent augmentation. Following prior work \cite{Zhu2020FreeLB:}, we use uniform distribution based perturbation at embedding level to produce adversarial examples during pre-training (task agnostic) and fine-tuning (task specific) stages. In addition, for our work, we use Gaussian noise based perturbation which actually results in a better performance.

\begin{table*}[]
\begin{tabular}{@{}ccc|ccc|ccc@{}}
\toprule
                        \multicolumn{3}{c}{\textsc{Mami}} & \multicolumn{3}{c}{\textsc{Fbhm}}         & \multicolumn{3}{c}{\textsc{HarMeme}} \\ \cline{1-9}
                        \textsc{Uniter}  & \textsc{VisualBERT} & \textsc{Rob+Resnet} & \textsc{Uniter} & \textsc{VisualBERT} & \textsc{Rob+Resnet} & \textsc{Uniter}   & \textsc{VisualBERT}  & \textsc{Rob+Resnet}  \\
               84.99   & 85.00      & \textbf{85.50}      & \textbf{63.53}  & 60.88      & 63.10      & \textbf{78.51}    & 75.57       & 73.67       \\
\bottomrule
\end{tabular}
\caption{Performance (macro-F1 scores) of the vanilla hate meme detection models.}
\label{tab:evaluation_results}
\end{table*}

\section{Experimental Setup}
In this section we discuss the overall experimental setup comprising a description of the hateful meme detection models, the training framework and the hyperparameters. 

\begin{table*}[]
\footnotesize
\begin{tabular}{@{}rrrr|rrr|rrr@{}}
\toprule
                       & \multicolumn{3}{c}{\textsc{Mami}} & \multicolumn{3}{c}{\textsc{Fbhm}}         & \multicolumn{3}{c}{\textsc{HarMeme}} \\ \cline{2-10}
                       & \textsc{Uniter}  & \textsc{VisualBERT} & \textsc{Rob+Resnet} & \textsc{Uniter} & \textsc{VisualBERT} & \textsc{Rob+Resnet} & \textsc{Uniter}   & \textsc{VisualBERT}  & \textsc{Rob+Resnet} \\
\texttt{Add}                   &    \ApplyGradient{0.19}    &    \ApplyGradient{0.46}    &    \ApplyGradient{0.11}    &    \ApplyGradient{0.13}    &    \ApplyGradient{-0.60}    &    \ApplyGradient{0.35}    &    \ApplyGradient{-1.48}    &    \ApplyGradient{-2.90}    &    \ApplyGradient{-0.17}\\
\texttt{Blur}              &    \ApplyGradient{1.29}    &    \ApplyGradient{1.54}    &    \ApplyGradient{0.52}    &    \ApplyGradient{-1.44}    &    \ApplyGradient{0.69}    &    \ApplyGradient{-0.19}    &    \ApplyGradient{9.99}    &    \ApplyGradient{9.42}    &    \ApplyGradient{5.09}\\
\texttt{SaltPepper-T}          &    \ApplyGradient{0.09}    &    \ApplyGradient{0.77}    &    \ApplyGradient{1.24}    &    \ApplyGradient{-1.64}    &    \ApplyGradient{0.43}    &    \ApplyGradient{-1.10}    &    \ApplyGradient{3.89}    &    \ApplyGradient{2.24}    &    \ApplyGradient{-2.84}\\
\texttt{SaltPepper-I-Low}      &    \ApplyGradient{2.29}    &    \ApplyGradient{3.62}    &    \ApplyGradient{6.10}    &    \ApplyGradient{0.34}    &    \ApplyGradient{1.19}    &    \ApplyGradient{1.61}    &    \ApplyGradient{4.89}    &    \ApplyGradient{3.70}    &    \ApplyGradient{4.52}\\
\texttt{SaltPepper-I-High}    &    \ApplyGradient{4.42}    &    \ApplyGradient{5.50}    &    \ApplyGradient{5.11}    &    \ApplyGradient{2.04}    &    \ApplyGradient{2.65}    &    \ApplyGradient{3.92}    &    \ApplyGradient{16.62}    &    \ApplyGradient{12.02}    &    \ApplyGradient{8.27}\\
\texttt{Newsprint}            &    \ApplyGradient{6.81}    &    \ApplyGradient{8.57}    &    \ApplyGradient{14.47}    &    \ApplyGradient{3.29}    &    \ApplyGradient{3.80}    &    \ApplyGradient{0.54}    &    \ApplyGradient{7.86}    &    \ApplyGradient{7.52}    &    \ApplyGradient{4.58}\\
\texttt{Spread-Low}         &    \ApplyGradient{1.19}    &    \ApplyGradient{1.46}    &    \ApplyGradient{1.21}    &    \ApplyGradient{0.52}    &    \ApplyGradient{1.82}    &    \ApplyGradient{2.34}    &    \ApplyGradient{13.39}    &    \ApplyGradient{11.54}    &    \ApplyGradient{16.56}\\
\texttt{Spread-High}         &    \ApplyGradient{5.14}    &    \ApplyGradient{4.20}    &    \ApplyGradient{8.78}    &    \ApplyGradient{2.70}    &    \ApplyGradient{6.65}    &    \ApplyGradient{3.49}    &    \ApplyGradient{27.66}    &    \ApplyGradient{28.20}    &    \ApplyGradient{25.44}\\
\texttt{Add+SaltPepper-I}       &    \ApplyGradient{5.30}    &    \ApplyGradient{6.41}    &    \ApplyGradient{16.47}    &    \ApplyGradient{1.40}    &    \ApplyGradient{3.09}    &    \ApplyGradient{3.15}    &    \ApplyGradient{6.61}    &    \ApplyGradient{5.36}    &    \ApplyGradient{4.98}\\

\addlinespace[2 mm]

\bf Average       &    \ApplyGradient{2.97}    &    \ApplyGradient{3.61}    &    \ApplyGradient{6.00}    &    \ApplyGradient{0.82}    &    \ApplyGradient{2.19}    &    \ApplyGradient{1.57}    &    \ApplyGradient{9.94}    &    \ApplyGradient{8.57}    &    \ApplyGradient{7.38}\\

\bottomrule
\end{tabular}
\caption{\% $\Delta$ in macro-F1 hate meme detection systems are exposed to different adversarially modified test inputs. The $+ve$ values (highlighted with gradient \colorbox{red}{RED}) indicate the amount of performance drop when an attack has been applied while $-ve$ values (highlighted with \colorbox{green}{GREEN}) indicate performance improvements over the baselines.}
\label{evaluation_results}
\end{table*}

\subsection{Hate meme detection models} 
We select a series of state-of-the-art hate meme detection models in order to test their vulnerability against adversarial attacks. To analyse the model vulnerability, we start with reproducing  results for hate meme detection task. These models are discussed in brief for better readability. 

\noindent\textsc{VisualBERT}: This \cite{https://doi.org/10.48550/arxiv.1908.03557} builds up on the \textsc{BERT} architecture \cite{devlin-etal-2019-bert} and exploits the attention layer to align the input meme text with its image regions which are generated from variety of object detectors. Task specific MLM is used for pretraning. Unlike the original implementation where token ids are persistent during fine-tuning, we use the Vilio\footnote{\url{https://github.com/Muennighoff/vilio}} framework where weights for token type are trained from scratch which has been shown to exhibit promising performance in hate meme detection task \cite{https://doi.org/10.48550/arxiv.2012.07788}. Likewise we also use an additional classification head with very high learning rate (500 times the original) and multi-sample dropout.

\noindent\textsc{Uniter}: Like \textsc{VisualBERT}, the UNiversal
Image-TExt Representation (\textsc{Uniter}) model \cite{chen2020uniter} builds on an early fusion approach and is pre-trained on large text-image datasets. Visual features extracted from faster R-CNN \cite{NIPS2015_14bfa6bb} and textual ones from word piece encodings \cite{https://doi.org/10.48550/arxiv.1609.08144} are combined through a transformer based architecture to align them in a shared embedding space. The model emphasizes on fine grained alignment between the modalities with conditional masking and optimal transport \cite{https://doi.org/10.48550/arxiv.1803.00567}. We update the activation functions and embedding calculations as noted in \cite{https://doi.org/10.48550/arxiv.2012.07788}. 

\noindent\textsc{Rob+Resnet}: Our proposal is to adopt a late fusion approach so that the individual contributions of two modalities are better captured. To this purpose we suggest separate extraction pipelines for the image and the text features. To extract the image features we use \textsc{Resnet} - a very deep residual learning framework for image feature generation \cite{https://doi.org/10.48550/arxiv.1512.03385}. To obtain the text representation, we use the popular \textsc{RoBERTa} \cite{https://doi.org/10.48550/arxiv.1907.11692} model. Finally, we concatenate the features from both the modalities and pass it through a feed-forward network having 128 hidden layers with ReLu activation and dropout = 0.2 to generate the predictions. We train the model for 30 epochs on Adam optimizer \cite{https://doi.org/10.48550/arxiv.1412.6980} with learning rate = $10^{-5}$ and weight decay = 0.1. 

\noindent \textsc{CounterMeasures}: In addition to the vanilla implementation of the above-mentioned models, we repeat our experiments in presence of the two countermeasures discussed in the previous section to combat the effect of attacks while keeping the performance. For \emph{VILLA}, we use three different variants. These include \textsc{VILLA-PT-UN} where adversarial training is only performed during the pre-training phase with uniform noise, \textsc{VILLA-FT-UN} and \textsc{VILLA-FT-GN} where additional training is done at the fine-tuning stage with uniform and Gaussian noise respectively. In all the variants, the default number of adversarial steps is kept at 3. Due to the requirement of huge computation, we limit ourselves to not doing Gaussian noise-based perturbation at the pre-training stage. For contrastive learning, the loss is calculated for each modality and later on added to the loss function of the hate meme classification model. We test this for $\tau = 0.5$ and call it \textsc{CL-Ind-0.5}. We ensemble both the countermeasures (call it \textsc{Ensemble}) by taking averages of their prediction probabilities. 

\subsection{Model training}
To study the robustness of hate meme detection models, performance must be evaluated before and after the application of adversarial attacks. Therefore we start with introducing the attacks on the test instances for each dataset. Consequently for each dataset, we generate 10 different poisoned test sets including the original one. 

\noindent\textit{Text pre-processing}: First, we extract the text from the memes. To this purpose, we use the paid service API from Google Vision called  GoogleOCR\footnote{\url{https://cloud.google.com/vision/docs/ocr}}. We choose a meme and apply the OCR model directly on it without performing any preprocessing as the OCR is already equipped with its own preprocessor component. For tokenization, we use BERT-base \cite{devlin-etal-2019-bert} as well as RoBERTa-base tokenizers \cite{https://doi.org/10.48550/arxiv.1907.11692}. 

\noindent\textit{Image pre-processing} Following \cite{https://doi.org/10.48550/arxiv.2012.07788}, we extract the image features using the detectron2 framework~\cite{wu2019detectron2} because it makes the training process faster without compromising the relevant content. To be precise, we use Faster R-CNN with ResNet-101, using object and attribute annotations from Visual Genome \cite{Anderson_2018_CVPR}. The pretrained model generates bottom-up attention features corresponding to salient image regions. Minimum and Maximum boxes are set to 36 for such region of interests.


\vspace{1mm}

\noindent
Both text and image features are then fed to the hate meme detection models in accordance to their architecture requirements.

\noindent\textit{Hyperparameters}: Apart from the \textsc{CounterMeasures}, for rest of the detection models, we keep the default hyperparameters as provided in \cite{https://doi.org/10.48550/arxiv.2012.07788} implementation. Since \emph{VILLA} need to be trained on adversarial examples generated from each input steps we tune the variable which is responsible for setting the limits of the number examples. We choose this variable as  $1,2,3,4,5,6,10$, and 100. Generation of adversarial examples also depends upon the type of noise, i.e., uniform or Guassian noise. In case of contrastive learning, we tune the temperature coefficient $\tau$ which is responsible for distinguishing positive and negative samples. We perform our experimentation with $\tau = 0.5$ as has been noted earlier.

\section{Results}

This section is divided into three parts. In the first part, we report the performance of vanilla hate meme detection models. In the second part, we report the change in the performance when the data points are adversarially altered. In the third part, we report the performance (re)gained due to the introduction of the countermeasures. Since two out of the three datasets are unbalanced, we use macro-F1 to measure performance all through. 

\subsection{Performance of the hate meme detection models}
In Table~\ref{tab:evaluation_results} we report the performance of the vanilla hate meme models in terms of macro-F1. All the models perform almost similarly with \textsc{Rob+Resnet} doing slightly better for the \textsc{Mami} dataset and the \textsc{Uniter} model doing slightly better for the other two datasets. Explicit significant tests (M-W U test) show that none of the models significantly outperform one another.

\subsection{Adversarial attacks} 
To detect vulnerability of the hate meme detection models against adversarial attacks, we replace test set instances with their attacked versions. Finally, each test set have nine different noisy variants which are used for the analysis below as well as for evaluating the proposed countermeasures in the next subsection. Table~\ref{evaluation_results} illustrates the vulnerability of the existing state-of-the art hate meme detection models with GoogleOCR text recognition model. Each value in the table represents the percentage difference between the macro F1 score before and after application of the attacks. Therefore, greater the positive value, the more severe is the effect of the attack on the model. Almost all models suffer a drop in performance across the different datasets and for different attack schemes. For the \textsc{Mami} dataset the \texttt{Newsprint} noise strongly affects all the models with \textsc{Rob+Resnet} seeing a drop of as high as $\sim14.5\%$. The \texttt{Spread-High} noise is next in line among the unimodal attacks. Naturally, the multimodal attack \texttt{Add+SaltPepper-I} also adversely affects all the models with \textsc{Rob+Resnet} again seeing the highest drop of $\sim16.5\%$. Compared to the other two datasets, for the \textsc{FBHM} dataset all the models suffer less from the attacks. Here, the \textsc{VisualBERT} model suffers the most for the \texttt{Spread-High} noise, the \textsc{Rob+Resnet} suffers the most for the \texttt{SaltPepper-I-High} noise and \textsc{Uniter} model suffers the most for the \texttt{Newsprint} noise. For some textual noise, the models also show an improvement in performance which might be attributed to the fact that the training dataset possibly already had some similar adversarial examples by default which allowed the models to learn their characteristics. For the \textsc{HarMeme} dataset we observe that all the models suffer the most. Out of all, the \texttt{Spread-High} noise deteriorates the performance of all the models the most with 27.7\%, 28.2\% and 25.4\% reduction in macro-F1 scores for  \textsc{Uniter}, \textsc{VisualBERT} and \textsc{Rob+Resnet} respectively. On average all models suffer a reduction in the macro-F1 scores due to the introduction of the adversarial attacks with image based attacks being always more severe than text based attacks. Among the image based attacks, \texttt{SaltPepper-I-High}, \texttt{Spread-High} and \texttt{Newsprint} noises affect the model performances the most. For all the \text{Add} attacks we show the results when \emph{LOVE} is embedded in the memes. The results when \emph{CARESS} is embedded are the same, as is expected, and hence not shown.

\begin{table*}[]
\resizebox{\textwidth}{!}{
\begin{tabular}{@{}rrrr|rrr|rrr@{}}
\toprule
                       & \multicolumn{3}{c}{\textsc{Mami}} & \multicolumn{3}{c}{\textsc{Fbhm}}         & \multicolumn{3}{c}{\textsc{HarMeme}} \\ \cline{2-10}
                        & \textsc{CL-Ind-0.5} &  \textsc{VILLA-FT-GN}  & \textsc{Ensemble} & \textsc{CL-Ind-0.5} &  \textsc{VILLA-FT-GN}  & \textsc{Ensemble} & \textsc{CL-Ind-0.5} &  \textsc{VILLA-FT-GN}  & \textsc{Ensemble} \\
\addlinespace[2 mm]
\texttt{Add}                  &    
(4.61,-4.21)$\downarrow$	& (2.08,-1.68)$\downarrow$ & (1.43,-1.03)$\downarrow$ &
(-1.08,0)$\uparrow$	& (-5.19,0)$\uparrow$ & \bf (-6.39,0)$\uparrow$ &
\bf (-1.04,0)$\uparrow$	& (4.42,-4.1)$\downarrow$ & (3.3,-2.98)$\downarrow$  \\

\texttt{Blur}               &    
(5.37,-6.07)$\downarrow$	& (1.47,-2.17)$\downarrow$ & (1.12,-1.82)$\downarrow$ &
(-1.6,0)$\uparrow$	& (-3.14,0)$\uparrow$ & \bf (-4.75,0)$\uparrow$  &
(-0.34,-10.81)$\uparrow$	& (-2.91,-8.24)$\uparrow$ & \bf (-3.5,-7.65)$\uparrow$  \\

\texttt{SaltPepper-T}            &    
(7.6,-7.1)$\downarrow$	& (3.18,-2.68)$\downarrow$ & (2.05,-1.55)$\downarrow$ &
(-2.25,0)$\uparrow$	& (-4.33,0)$\uparrow$ & \bf (-4.35,0)$\uparrow$   &
(0.53,-5.58)$\downarrow$	& (-2.03,-3.02)$\uparrow$ & \bf (-2.56,-2.49)$\uparrow$\\

\texttt{SaltPepper-I-Low}       &    
(7.8,-9.5)$\downarrow$	& (5.7,-7.4)$\downarrow$ & (2.23,-3.93)$\downarrow$ &
(-2,0)$\uparrow$	& \bf (-2.44,0)$\uparrow$ & (-2.07,0)$\uparrow$ &
(3.66,-9.71)$\downarrow$	& (1.47,-7.52)$\downarrow$ & (1.59,-7.64)$\downarrow$  \\

\texttt{SaltPepper-I-High}     &    
(9.32,-13.15)$\downarrow$	& (3.73,-7.56)$\downarrow$ & (2.38,-6.21)$\downarrow$ &
(-0.08,-1.89)$\uparrow$	& (-1.9,-0.07)$\uparrow$ & \bf (-1.91,-0.06)$\uparrow$ &
(-1.65,-16.13)$\uparrow$	& (-1.81,-15.97)$\uparrow$ & \bf (-3.66,-14.12)$\uparrow$ \\

\texttt{Newsprint}             &    
(11.12,-17.34)$\downarrow$	& (4.81,-11.03)$\downarrow$ & (4.93,-11.15)$\downarrow$ &
(4.64,-7.86)$\downarrow$	& (0.26,-3.48)$\downarrow$ & (0.25,-3.47)$\downarrow$ &
(2.56,-11.58)$\downarrow$	& (1.06,-10.08)$\downarrow$ & (2.05,-11.07)$\downarrow$ \\

\texttt{Spread-Low}         &    
(3.4,-4)$\downarrow$	& (2.05,-2.65)$\downarrow$ & (1.02,-1.62)$\downarrow$ &
(-2.32,0)$\uparrow$	& (-3.67,0)$\uparrow$ & \bf (-4.16,0)$\uparrow$ & 
(1.58,-16.13)$\downarrow$	& \bf (-1.83,-12.72)$\uparrow$ & (-1.79,-12.76)$\uparrow$ \\

\texttt{Spread-High}           &    
(4.72,-9.27)$\downarrow$	& (-0.25,-4.3)$\uparrow$ & (0.48,-5.03)$\downarrow$ &
(1.92,-4.55)$\downarrow$	& \bf (-1.76,-0.87)$\uparrow$ & (-0.22,-2.41)$\uparrow$ &
(3.32,-32.14)$\downarrow$	& (-5.03,-23.79)$\uparrow$ & \bf (-5.08,-23.74)$\uparrow$ \\

\texttt{Add+SaltPepper-I}        &    
(16.06,-20.77)$\downarrow$	& (6.21,-10.92)$\downarrow$ & (4.82,-9.53)$\downarrow$  &
(3.36,-4.69)$\downarrow$	& (1.8,-3.13)$\downarrow$ & (0.99,-2.32)$\downarrow$ &
(3.7,-11.47)$\downarrow$	& (2.41,-10.18)$\downarrow$ & (2.03,-9.8)$\downarrow$ \\
\addlinespace[2 mm]
\bf Average &    
(7.78,-10.75)$\downarrow$	& (3.22,-6.19)$\downarrow$ & (2.28,-5.25)$\downarrow$  &
(0.06,-0.88)$\downarrow$	& (-2.27,0)$\uparrow$ & \bf (-2.52,0)$\uparrow$ &
(1.37,-11.31)$\downarrow$	& (-0.48,-9.46)$\uparrow$ & \bf (-0.85,-9.09)$\uparrow$ \\
\bottomrule
\end{tabular}
}
    \caption{Performance of best version of each countermeasure over the \textsc{Uniter} model. Each cell in the table is a $(x,y)$ tuple as defined in the text. $\downarrow$ and $\uparrow$ respectively indicate whether a countermeasure worsens or improves the attacked model's performance in terms of the macro-F1 score. We bold out the best countermeasures (only improvement) on each attack across the datasets.}
\label{evaluation_results_diff}
\end{table*}


\subsection{Countermeasures}
In this section we present the results of our countermeasure approaches. We choose \textsc{Uniter} as the model since it showed slightly better performance in two out of the three datasets (observations from the other models show similar trends). We compare the adverserially attacked model performance in presence of the countermeasure techniques with those in the absence of these techniques. Precisely, we call the percentage difference in the macro-F1 score of the adverserially attacked model in absence and presence of countermeasures as $x$. Therefore a higher negative value of $x$ indicates that a particular countermeasure is more effective in reducing the adverse effect of the attack on the model. We also measure how close the countermeasure brings the model performance to the original unattacked model. Thus, we compute the percentage difference in the macro-F1 score of the unattacked model and the attacked model with the countermeasure and call it $y$. The expected value of $y$ is 0, i.e., after the application of the countermeasure the model performance should be equivalent to the performance of the unattacked model (which was also the main purpose of introducing the countermeasure in the first place). In general, a lower negative value of $y$ should be better.

\noindent\textbf{Key results and observations}: Table~\ref{evaluation_results_diff} presents the results of this experiment. Each entry in the table shows the $(x,y)$ values for a particular attack and a particular countermeasure. We show the performance of \textsc{CL-Ind-0.5} and \textsc{VILLA-FT-GN} since we find them to be the best among all the variants. Further, to unfold the best of these two countermeasures we also present an \textsc{Ensemble} of both of them by considering average of their prediction probabilities. Overall, we make the following observations.
\begin{compactitem}
    \item For the \textsc{Fbhm} and the \textsc{HarMeme} datasets, the countermeaures are effective for multiple of the attack strategies. For both these datasets, on average, the \textsc{Ensemble} countermeasure performs the best followed by \textsc{VILLA-FT-GN} and \textsc{CL-Ind-0.5} countermeasures in that order.
    \item For the \textsc{Fbhm} dataset the highest improvement in case of text based attacks is obtained for \texttt{Add} using the \textsc{Ensemble} based countermeasure. The performance in presence of the countermeasure improves by 6.39\% over the performance in absence of the countermeasure. In fact, in this case, the difference between the performance of the attacked model with countermeasure and the unattacked model is 0 as expected. The two other text based attacks -- \texttt{Blur} and \texttt{SaltPepper-T} -- also see large benefits in presence of the \textsc{Ensemble} based countermeasure. Among the image based attacks, the \textsc{VILLA-FT-GN} and the \textsc{Ensemble} based countermeasures are most effective for the \texttt{SaltPepper-I-Low} and the \texttt{SaltPepper-I-High} attacks respectively. For \texttt{Spread-Low} attack, \textsc{Ensemble} based countermeasure is most effective and for \texttt{Spread-High}, \textsc{VILLA-FT-GN} is most effective.  
    \item For the \textsc{HarMeme} dataset, the \textsc{CL-Ind-0.5} countermeasure does best for the \texttt{Add} attack. For the other two text based attacks the \textsc{Ensemble} based countermeasure work the best. In the image based attacks, the \textsc{Ensemble} based countermeasure once again works the best for the \texttt{SaltPepper-I-High} and the \texttt{Spread-High} attacks. \textsc{VILLA-FT-GN} is most effective for the \texttt{Spread-Low} attack.
    \item For both the \textsc{Fbhm} and \textsc{HarMeme} datasets none of the countermeasures work for certain attacks like \texttt{Newsprint} and \texttt{Add+SaltPepper-I}\footnote{$x$ becomes positive in these cases which means the introduction of the countermeasure worsens the performance} and is a scope for future improvement.
    \item Finally, none of the countermeasures seem to work for the \textsc{Mami} dataset. On a deeper investigation we find that the resolution of the original images for this dataset is already very bad; the average \emph{bit depth} for this dataset is 4.30 compared to 43.90 and 9.54 for the \textsc{HarMeme} and the \textsc{Fbhm} datasets. This is possibly the reason why the introduction of the adversarial examples in the training phase does not benefit the models.  
\end{compactitem}

\begin{figure}
  \begin{center}
  \resizebox{0.38\textwidth}{!}{%
\includegraphics[]{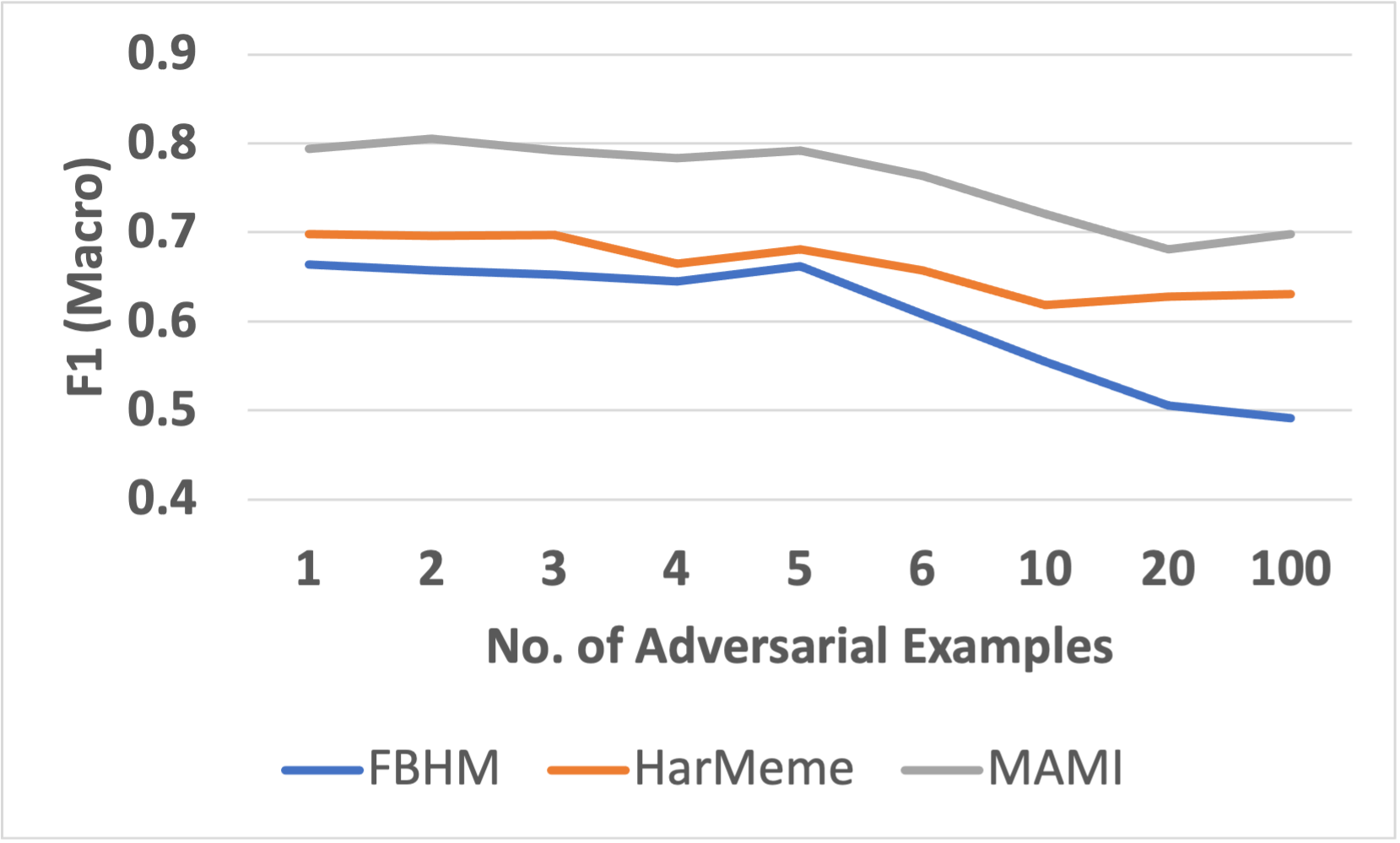}
}
        \end{center}
        \caption{Average macro-F1 over all the attacks with increase in the adversarial examples used during fine-tuning process of \emph{VILLA} model.}

        \label{advexample}
\end{figure}

\noindent\textbf{Impact of the number of adversarial examples}: We further observe that the number of adversarial examples added in the training set has a positive impact only up to a certain extent. During the fine-tuning process of \emph{VILLA}, for each training input sample we vary the number of adversarial examples that we add and estimate the performance after each addition. is used to generate the multiple adversarial examples. All other parameter settings remain the same. Figure~\ref{advexample} illustrates that the inclusion of up to five adversarial examples is tolerable by the system after which it steadily deteriorates.

\if{0}\begin{figure}
  \begin{center}
  \resizebox{0.45\textwidth}{!}{%
\includegraphics[]{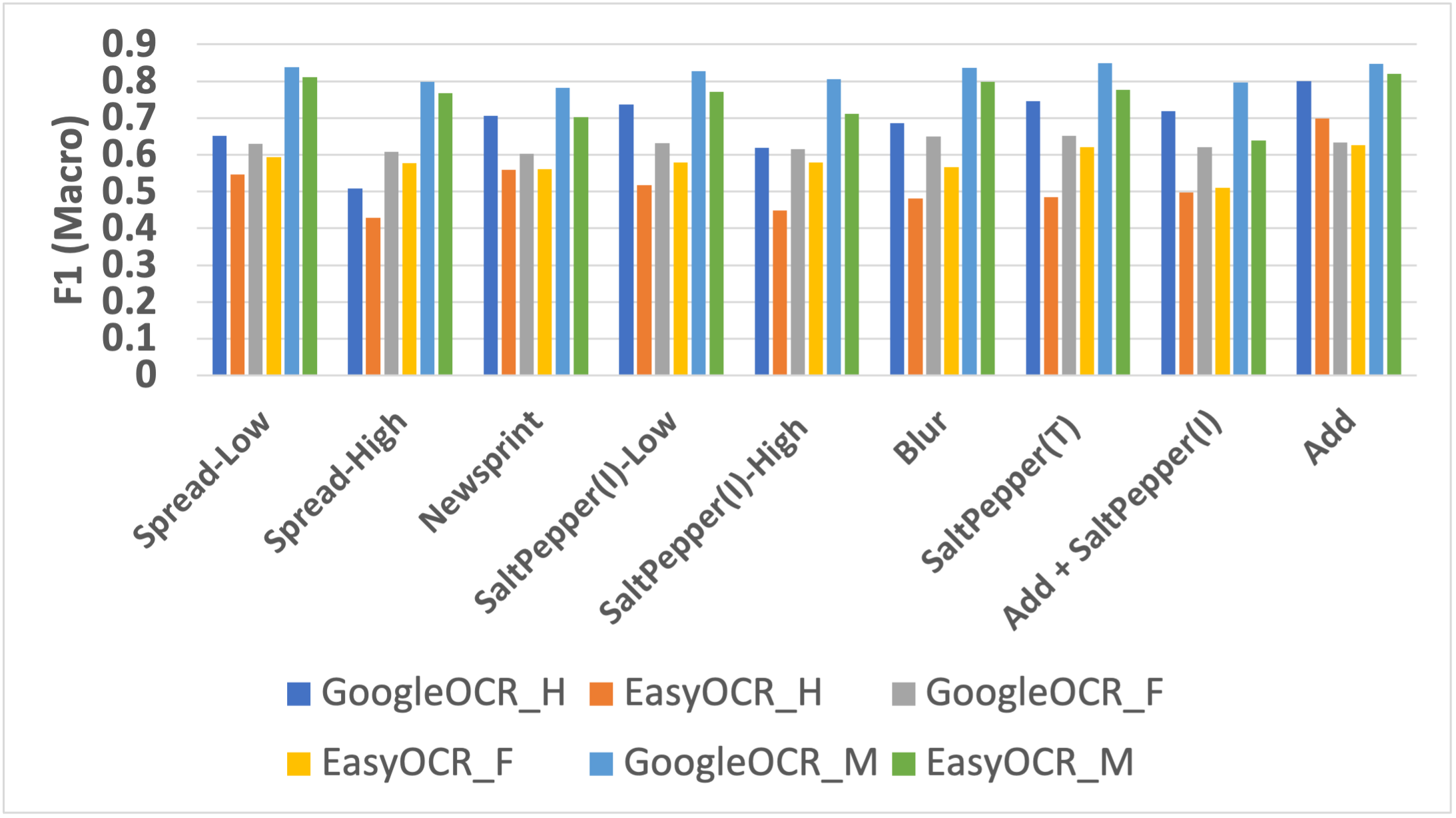}
}
        \end{center}
        \caption{OCRs impact on the robustness of Hateful Meme Detection.\ps{change images to pdf, also draw them in matplotlib}}

        \label{ocr}
\end{figure}\fi

\if{0}\section{Discussion}
Finally, we present analysis which focus on fluctuations in model behavior with respect to certain aspect. We also present limitations as well as challenges of our analysis which definitely open room for further research. 

\paragraph{OCRs on Memes} Availability of meme text in addition to meme is hardly a case in real time setting. Therefore, we must apply OCRs to extract text from the memes which further needed to extract textual features. However, no OCR is 100 percent confident about its output. Therefore, such noise can make the models highly vulnerable. In this paper, we compare the already discussed paid variant of OCR called googleOCR with a freely available and popularly used API called easyOCR\footnote{\url{https://github.com/JaidedAI/EasyOCR}} which recognize text based on combination of \emph{Resnet} and  \emph{LSTM} models. We find that easyOCR is consistently poorer than GoogleOCR across all the proposed attacks (see Figure~\ref{ocr}). \fi


\section{Conclusion and Outlook}
In this study, we systematically analyse vulnerability of multi-modal hate meme detection systems when they are attacked with partial model knowledge based adversaries. We audited widely popular visual linguistic models for a number of text and image based attacks and observed that there is a steep performance decrease in performance. Overall the image based attacks are found to be more severe. In order to counter such attacks we proposed two different methods -- contrastive learning and adversarial training and found that an ensemble of these two methods work well for a large majority of attacks for two of the three datasets. 


\noindent\textbf{Future directions}: A natural follow up of this work is to develop countermeasures that would work for very severe adversarial attacks like \texttt{Newsprint} and \texttt{Add+SaltPepper-I}. Another direction would be to identify countermeasure approaches for datasets where the original images are themselves of very low quality (e.g., the \textsc{Mami} dataset in our study). Last but not the least, the set of our attack schemes are far from exhaustive and more variants need to be tried in future.

\section*{Acknowledgments}
This work was supported by CATALPA - Center of Advanced Technology for Assisted Learning and Predictive Analytics of the FernUniversität in Hagen, Germany.

\bibliographystyle{ACM-Reference-Format}
\bibliography{sample-base}

\appendix
\section{Ethical Consideration}
\subsection{Unintentional bias} Any biases found in our study are unintentional, and we do not intend to do harm to any group or individual. We note that determining whether a meme is harmful can be subjective, and thus it is inevitable that there would be biases in model training process and its analysis. 
\subsection{Code misuse} We intend to publish our code base in order to replicate our study for further research. However, it can also be exploited to generate attacks in order to confuse the algorithms deployed by social media. Therefore,  human moderation in addition to algorithmic detection is needed in order to ensure that this does not occur.

\subsection{Carbon emission} We carried out most of our experiments on GPUs to analyse their robustness and to generate counter-measure models. Our experiments were conducted using a private infrastructure, which has a carbon efficiency of 0.432 kgCO$_2$eq/kWh. A cumulative of 300 hours of computation was performed on hardware of type RTX 2080 Ti (TDP of 250W). Total emissions are estimated to be 32.4 kgCO$_2$eq of which 0\% was directly offset. Estimations were conducted using the \href{https://mlco2.github.io/impact#compute}{Machine Learning Impact calculator} presented in \cite{lacoste2019quantifying}. 
\end{document}